%% file: main.tex
\documentclass[runningheads]{llncs}
\usepackage[T1]{fontenc}
\usepackage{graphicx}
 \usepackage{xcolor}
\usepackage{subcaption}
\usepackage{float}
\usepackage{soul}
\usepackage{tabularx} 
\usepackage{amsmath}

\usepackage{rotating}
\usepackage{graphicx}

\begin{document}

\title{Validating Virtual Reality for Studying Multimodal Human-Robot Interaction in Socially Aware Robot Navigation}

\titlerunning{Validating VR for Studying Multimodal HRI in Socially Aware Navigation} 

\author{Hariharan Arunachalam\inst{1}
\orcidID{0009-0009-9257-4709} 
\and
Phani Teja Singamaneni\inst{2}
\orcidID{0000-0003-4513-8954} 
\and
Rachid Alami\inst{1}
\orcidID{0000-0002-9558-8163}
}
\authorrunning{H. Arunachalam et al.}

\institute{LAAS-CNRS, Universite de Toulouse, France 
\\ 
\email{harunachal@laas.fr , rachid.alami@laas.fr} \and
Inria Nancy, France \\
\email{phani-teja.singamaneni@inria.fr}
}

\maketitle              
\begin{abstract}

Virtual Reality (VR) offers a flexible and controllable platform for studying human-robot interaction. Prior work has explored VR for socially aware robot navigation. However, whether VR captures the multimodal interaction dynamics observed in real-world human-robot co-navigation remains insufficiently understood. 
In this work, we present a VR prototype and evaluate its suitability for studying multimodal human-robot interaction (HRI) in socially aware navigation. Specifically, we investigate whether VR preserves the multimodal interaction dynamics observed in real-world human-robot co-navigation. We conducted a within-subjects study (N = 21) in which participants interacted with a PR2 mobile manipulator robot in both a motion capture equipped arena and its virtual replica in an immersive VR environment. Two common co-navigation scenarios were examined : orthogonal crossing and pass-by interactions. Participants evaluated the robot's perceived social awareness and interaction comfort, while trajectory and head-orientation data were analysed to examine behavioral responses during the interaction. Our results show that participants perceive the robot’s socially aware navigation similarly in VR and in the real world. Furthermore, VR captures human interaction behaviors in ways consistent with real-world observations. These findings suggest that VR can be a reliable and flexible platform for studying richer multimodal behaviors in social navigation and HRI.

\end{abstract}

\section{Introduction}
\input{introduction}

\section{Related Works}
\input{related_works}

\section{System Architecture}
\input{methodology}

\section{User Study}
\input{experiments_and_user_survey}

\input{conclusion}

\bibliographystyle{splncs04}  
\bibliography{refs}

\end{document}

%% file: introduction.tex
\label{sec:introduction}
Socially-aware Robot Navigation (SRN) \cite{singamaneni2024survey,kruse2013human} focuses on planning the motion of robots so that they behave in a socially compliant manner when sharing space with humans. Beyond simple collision avoidance, robots must manage conflicts and interactions in a way that is legible, predictable, and comfortable for nearby people. Traditionally, humans in this planning problem are modeled as two-dimensional entities whose behavior is inferred primarily from their motion trajectories \cite{singamaneni2024survey}. Under this formulation, HRI is largely reduced to geometric reasoning, where humans are treated similar to dynamic obstacles whose future motion must be predicted and avoided.

However, human social perception relies on a much richer set of cues. People interpret the intentions and behaviors of others through multiple modalities, including body orientation, gaze direction, gestures, interpersonal distance, and contextual information \cite{basili2012inferring}. These signals play an important role in coordinating movements and negotiating shared spaces. As a result, modeling humans solely through two-dimensional motion trajectories provides only a limited representation of the factors that shape human-robot interaction.

Studying these multimodal aspects of social navigation in the real world can be challenging \cite{buisan:hal-04240696} due to limited experimental control, safety considerations, and the difficulty of systematically manipulating environmental and interaction variables. Virtual Reality (VR) provides an alternative platform, offering a controllable and reproducible environment in which complex human-robot interactions can be studied while capturing rich behavioral signals \cite{inamura2021sigverse} and high fidelity data.

While VR has been successfully used for robotics and HRI experiments, 
existing VR-based SRN studies primarily focus on planar human trajectories and interaction distances. These approaches rarely capture or validate additional behavioral signals such as head orientation, gaze direction, or body posture that are known to play a critical role in human motion prediction and coordination. Consequently, it remains unclear whether VR environments can reliably reproduce the multimodal behavioral cues observed in real-world human navigation. Addressing this question is important for validating VR as a platform for studying multimodal HRI in SRN. 

We address this question by developing a VR prototype that replicates a real-world human-robot co-navigation environment with a PR2 robot, and conduct a within-subjects study comparing interactions in real world and VR. In this paper, multimodal refers specifically to locomotion trajectories and head-orientation cues during co-navigation. By analyzing participant perceptions, their motion trajectories and head-orientations, we \textit{evaluate whether VR can reproduce interaction dynamics similar to those observed in the real world}, and serve as a valid platform for studying multimodal SRN. 
The results provide insights into the suitability of VR as a tool for studying richer multimodal behaviors in SRN. 

The main contributions of this work can be listed as follows:

1. We developed a VR prototype for 
replicating and studying SRN scenarios.

2. We validated if VR can capture multimodal interactions and serve as a platform for conducting preliminary user-studies in SRN.

3. We conducted a comparative user study and provided qualitative and quantitative analyses that show significant similarities between VR and real-world. \\ 

The remainder of this paper is organized as follows. Section~\ref{sec:related_works} reviews the related work relevant to this study. Section~\ref{sec:framework} presents the VR system architecture. Section~\ref{sec:user_study} describes the user study, including the scenario design and the measurement methods used in the evaluation. Section~\ref{sec:results} reports the results and provides both subjective and objective analyses. Section~\ref{sec:limitations} discusses the limitations of the current work. Finally, Section~\ref{sec:conclusion} concludes the paper and outlines potential directions for future research.

%% file: related_works.tex
\label{sec:related_works}
\subsection{Virtual Reality in Robotics}
Virtual Reality provides immersive, controllable environments that enable safe and reproducible human-robot interaction studies \cite{dianatfar2021review}. VR has been applied to teleoperation \cite{zuchowicz_leveraging_2025}, robot behavior prototyping \cite {alves_developing_2022}, and experimental studies of human behavior \cite{fratczak_understanding_2019,liu_understanding_2017}. Compared to real world experiments \cite{stratton:hal-05463565,kruse2014evaluating}, VR allows researchers to systematically manipulate environmental conditions, robot behaviors, and interaction scenarios while ensuring participant safety, experimental repeatability, and cost efficiency. Prior research indicates that VR can effectively support user studies and behavioral data collection, although perceptual differences compared to the real world should be considered \cite{leblong_wheelchair_2024,tsoi_influence_2024}. 

\subsection{Simulation and VR for Socially Aware Robot Navigation} 
Virtual environments and simulators are also widely used to evaluate SRN methods. These approaches can be broadly categorized according to their purpose and the level of human realism they aim to capture. First, many works rely on synthetic human agents to evaluate SRN in controlled benchmarking environments. For example, simulators such as 
MengeROS \cite{aroor2017mengeros}, CrowdBot \cite{grzeskowiak2021crowd} provide reproducible environments where robot interacts with a simulated pedestrian generated from simplified behavioral models. These platforms are particularly useful for debugging SRN and benchmarking performance at scale, although the behavioral diversity of simulated humans may remain limited.

A second category focuses on human decision-driver simulation, where simulated humans are controlled by complex predefined behavior models designed to reproduce different human strategies during co-navigation. For example, InHuS \cite{favier2022intelligent} and IMHuS \cite{hauterville2022imhus} introduced simulation frameworks where human agents are controlled through different behavior models along with navigation planners, enabling different human behaviors based on the context during interaction, allowing systematic study of SRN.
HuNavSim \cite{escudero2025hunavsim} proposes simulator independent framework with similar customizable human behavior models for SRN studies.

A third class of approaches involves human-controlled agents in simulation, typically non-immersive interfaces or bird's-eye views \cite{tsoi2021approach}. These systems allow human participants to control (teleoperate) virtual pedestrians interacting with the robot. However, the indirect interaction interface may alter natural navigation behavior.

These simulation-based approaches provide valuable tools for algorithm development and controlled experimentation. However, they often abstract away important aspects of real human behavior. In this context, immersive virtual reality \cite{inamura2021sigverse} offers a complementary approach, enabling human-in-the-loop studies where participants perceive and react to robots while still benefiting from experimental control.

Several studies have used VR to investigate SRN \cite{li_comparing_2019,lynch_effect_2018,leblong_wheelchair_2024,grzeskowiak_toward_2020}. For example, prior work has examined how humans perceive robot and interpersonal distance regulation \cite{li_comparing_2019}, revealing the perception biases in VR compared to real world setting. The study in \cite{grzeskowiak_toward_2020} compare human velocity in response to robot's navigation in both VR and real world environment, demonstrating that VR can be an effective tool for studying social navigation. 
Recent work in \cite{zhang_sean-vr_2023} develops a VR system for capturing facial features \cite{zhang_predicting_2025} from humans during navigation, enabling estimation of the user's subjective perception about the robot.  Together these works demonstrate the potential of VR for controlled SRN experimentation.  While these works provide valuable insights, they do not explicitly assess whether VR can faithfully reproduce multimodal navigation behaviors observed in real world. Our work addresses this gap by directly comparing matched real-world and virtual co-navigation experiments. Specifically, we evaluate whether VR preserves both locomotion trajectories and head-orientation patterns, providing empirical evidence for its suitability as a platform for multimodal SRN studies.

\subsection{Multimodal Cues in Social Navigation}
During navigation in shared environments, humans rely on multiple behavioral cues to anticipate the motion of others and coordinate their movements \cite{basili2012inferring}. Humans often look towards their intended trajectory, providing predictive cues for collision avoidance and goal estimation \cite{holman_watch_2021,unhelkar_human-robot_2015}. While eye gaze measurements can be noisy \cite{fang_embodied_2015}, head orientation serves as a coarse, reliable proxy for attention and intention \cite{kar_review_2017,higgins_head_nodate}. Therefore, in this work, we capture and analyze human head orientation while generating similar cues from the robot.

Apart from head orientations, body posture and orientation \cite{zhang2021multimodal} convey intended motion direction and coordination strategies in shared spaces. Waist \cite{yamauchi2025waist}, and specifically Pelvis \cite{farkhatdinov2017anticipatory} orientations also serve as an anticipatory signal for estimating human navigation.
These cues are essential for understanding human navigation and designing SRN behaviors. This motivates our investigation into multimodal human behavior for SRN in this work.

%% file: methodology.tex
\begin{figure}[ht]
    \centering
    \includegraphics[width=\linewidth]{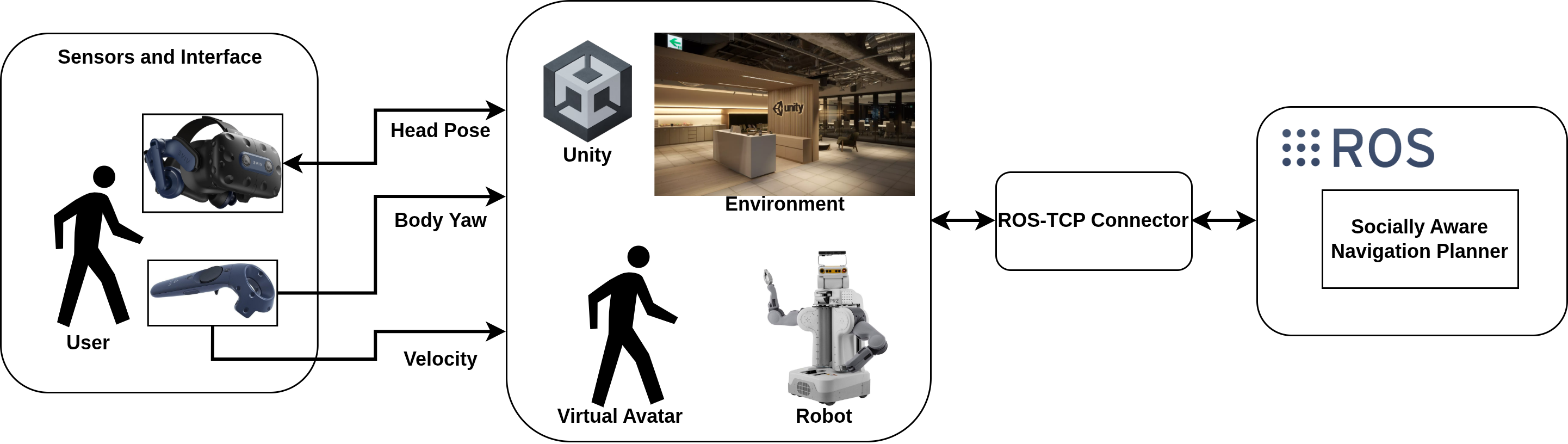}
    \caption{Architecture of the VR prototype that includes four main components, Sensors and Interface, Unity, ROS-TCP-Connector, and ROS}
    \label{fig:isonav_architecture}
\end{figure}

\label{sec:framework}
This section introduces the VR framework developed for this study and its main components. Fig.\ref{fig:isonav_architecture} illustrates the architecture of the VR simulation pipeline developed. 

\noindent It primarily consists of four main components, detailed below: 
\begin{enumerate}
    \item \textbf{Sensors and Interface}: This component allows participants to interact through a virtual avatar of the human which is controlled by the user in the simulation. It enables users to visualize the environment through a virtual camera attached to the virtual avatar's head, providing a first-person perspective of navigation while capturing their motion and behavioral cues such as head orientation of the user. Through this interface, the human participant controls a virtual avatar inside the simulation and interacts with the robot and environment.
    
    The current setup includes a HTC vive pro 2 headset and a handheld controller.
    \begin{itemize}
        \item \textit{Controlling the Virtual avatar}: 
        The user controls the yaw of the virtual avatar through the yaw of the handheld controller, while forward velocity is controlled through the pitch of the controller. The pitch is linearly mapped with the velocity of the virtual avatar, with a maximum speed of 1.5~m/s at maximum controller pitch.
        The virtual avatar is also animated to simulate realistic walking based on this velocity. This provides a more immersive and natural navigation experience for the participant.

        \item \textit{Controlling the Head Pose}: 
        While wearing the VR headset, the user experiences the virtual environment and their head orientation relative to a fixed reference frame is captured and translated into the head orientation of virtual avatar in the simulation, capturing the head orientation.
    \end{itemize}
    
    \item \textbf{Unity}: 
    Unity is used to simulate the environment, the robot, and the virtual avatar controlled by the participant. It provides controllable and reproducible environments suitable for navigation experiments, along with plugins that enable seamless control of both the robot and the virtual avatar.
    Unity also supports integration of various sensors such as cameras and LiDAR, which are important for robot localization and navigation. For the virtual avatar, we use Microsoft's Rocketbox\footnote{\url{https://github.com/microsoft/Microsoft-Rocketbox}} which contains animatable avatars that are compatible with Unity.
    
    \item \textbf{ROS-TCP Connector}:
    The framework uses the official ROS-TCP-Connector\footnote{\url{https://github.com/Unity-Technologies/ROS-TCP-Connector}} provided by Unity to enable communication between Unity and a ROS-based system. This connector allows commands to be sent to the robot controller while simultaneously using and acquiring data from the simulation.
    
    \item \textbf{ROS}: 
    Robot control is implemented using ROS (Robot Operating System). The robot's motion and head orientation are controlled through ROS based on information received from the simulation.
\end{enumerate}

%% file: experiments_and_user_survey.tex
\label{sec:user_study}
To validate our system, we conducted a user study after getting approval from our university ethics board. 21 participants were recruited for this within-subjects study (6 female, 15 male) with mean age of 27.7 (SD = 4.4). 
10 of them had some prior experience with VR, and 15 had some prior experience interacting with real robots.

\subsection{Scenario Design and Experiments}
\begin{figure}[ht]
    \centering
    \begin{subfigure}[b]{0.4\linewidth}
        \includegraphics[width=\linewidth]{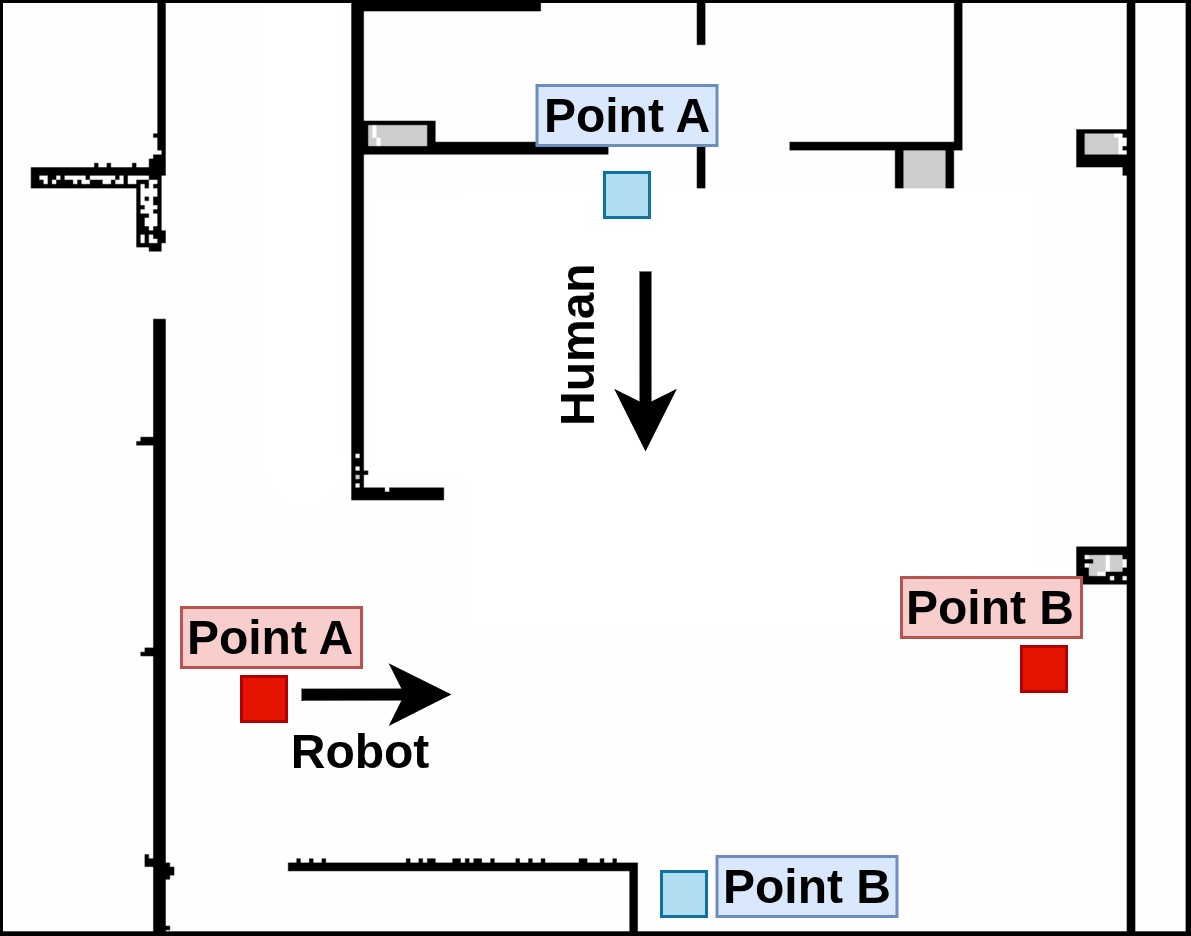}
        \caption{Orthogonal Crossing}
        \label{fig:orthogonal_scenario}
    \end{subfigure}
        \begin{subfigure}[b]{0.4\linewidth}
        \includegraphics[width=\linewidth]{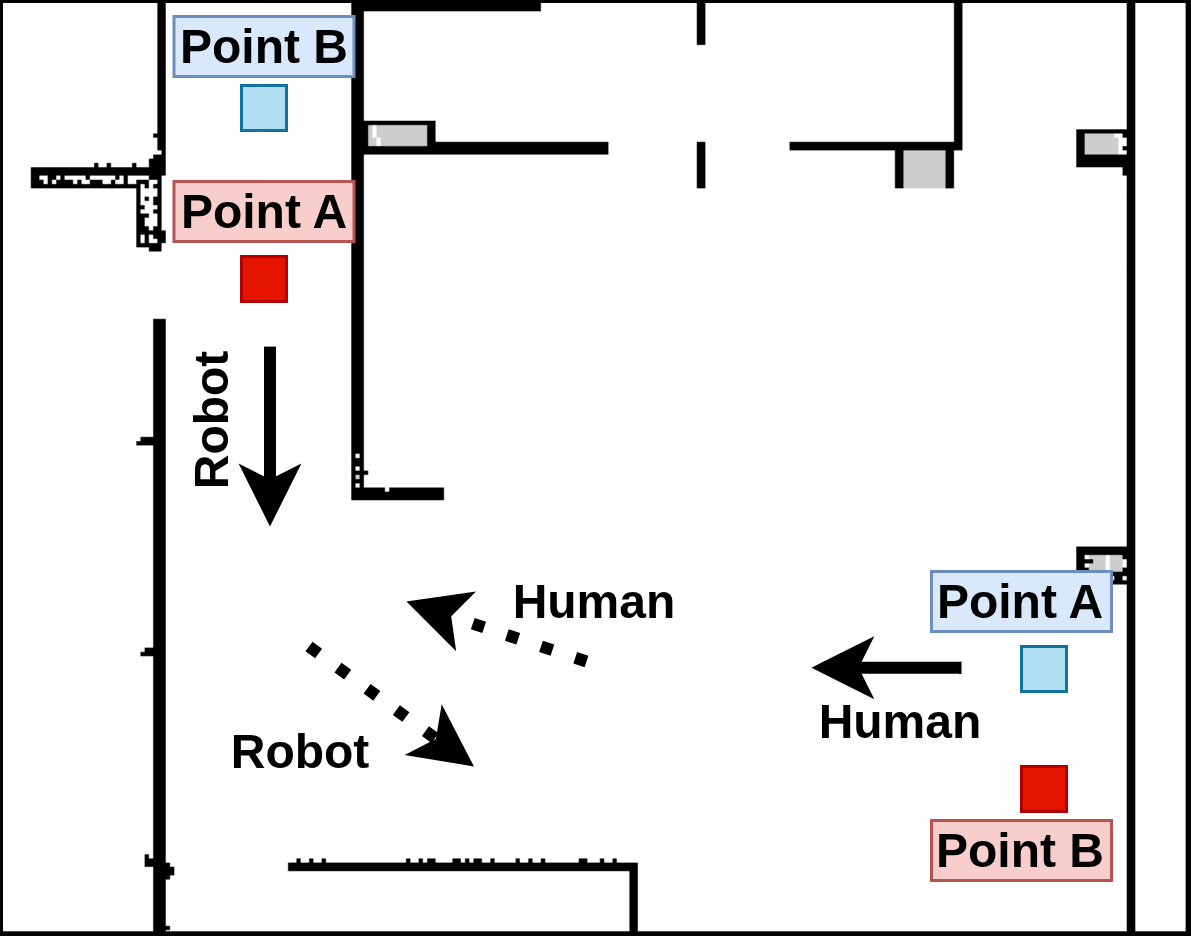}
        \caption{Passing By}
        \label{fig:pass_by_scenario}
    \end{subfigure}
    \caption{Navigation scenarios developed for the study. Blue indicates human goal positions, and Red indicates robot goal positions.  }
    \label{fig:scenarios_developed_for_collecting_the_data}
\end{figure}
\noindent To capture and study human behavior during co-navigation with a robot, two navigation scenarios were created: 
\vspace{-0.2cm}
\begin{itemize}
    \item \textbf{Orthogonal Crossing (Ortho)}: As shown in Fig. \ref{fig:orthogonal_scenario}, the human and robot cross each other in an orthogonal intersection while navigating to their respective goals, requiring coordinated path planning and timely motion adjustments to avoid collisions. 
    \item \textbf{Passing By (Passby)}: In this scenario, the human and robot pass by each other during navigation, as shown in Fig. \ref{fig:pass_by_scenario} requiring head-on collision avoidance decisions. 
\end{itemize}

\begin{figure}
    \centering
    \begin{subfigure}[t]{0.25\linewidth} 
        \centering
        \includegraphics[width=\linewidth,keepaspectratio]{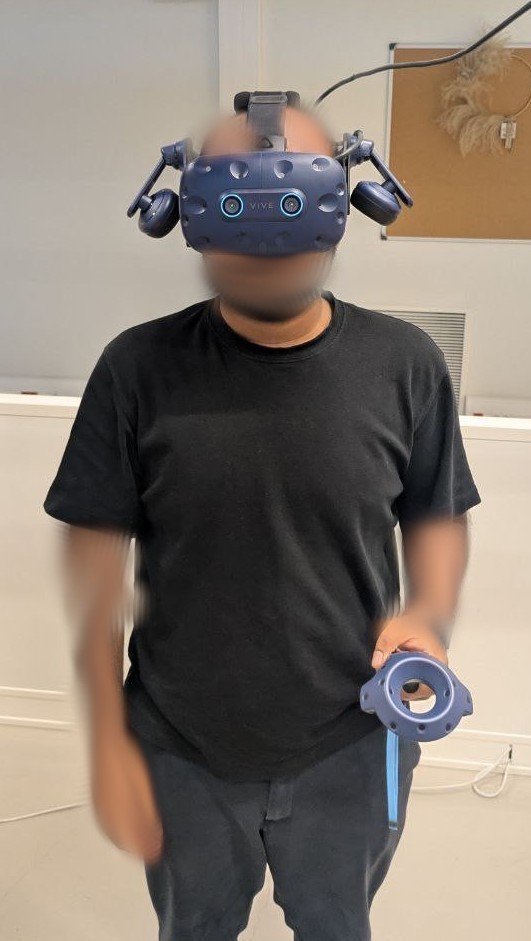}
        \subcaption{User with VR interface.}
        \label{fig:user_with_vr_interface}
    \end{subfigure}%
    \hspace{0.03\linewidth} 
    \begin{subfigure}[t]{0.68\linewidth} 
        \centering
        \includegraphics[width=\linewidth,keepaspectratio]{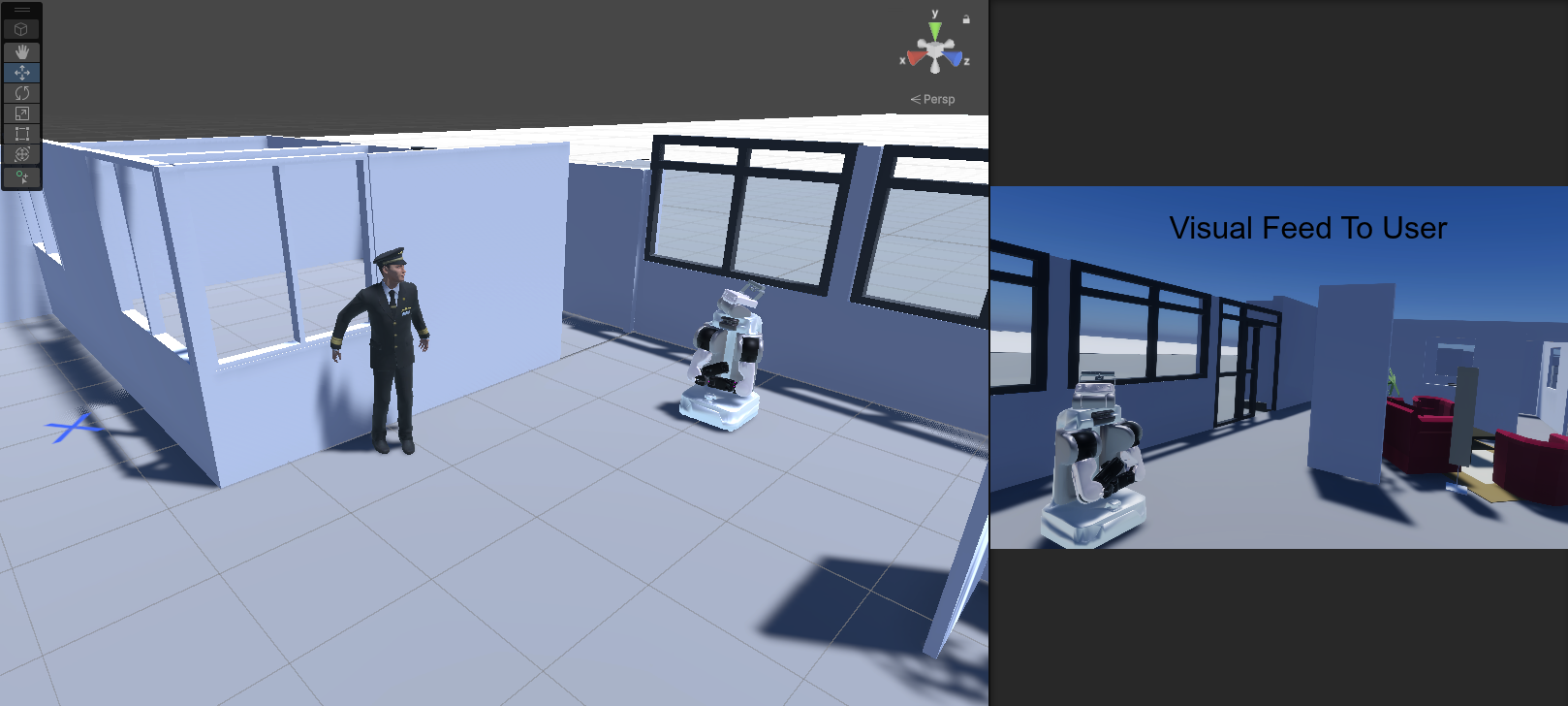}
        \subcaption{VR interaction and visual feed.}
        \label{fig:vr_image}
    \end{subfigure}

    \vspace{0.5cm}

    \begin{subfigure}[t]{0.25\linewidth} 
        \centering
        \includegraphics[width=\linewidth,keepaspectratio]{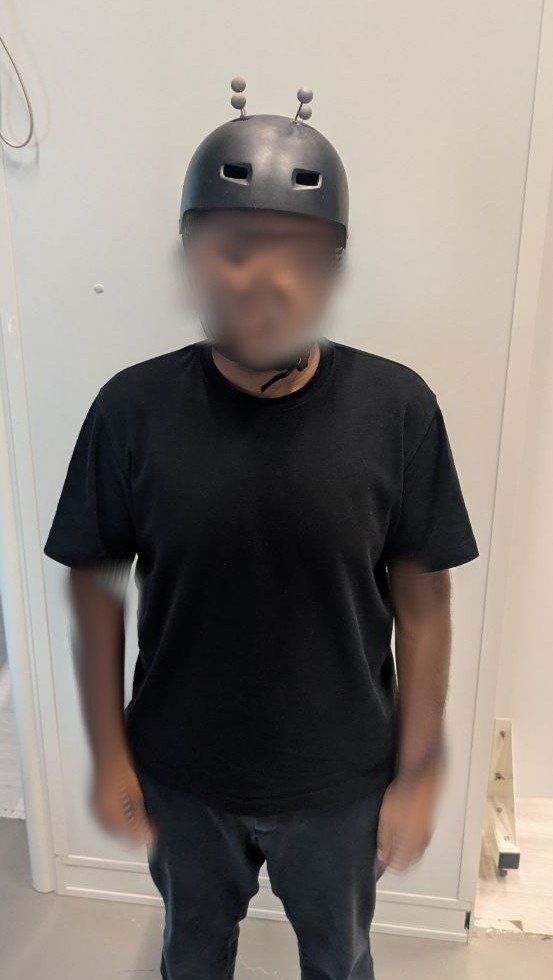}
        \subcaption{Motion capture helmet}
        \label{fig:helmet_and_body_markers}
    \end{subfigure}%
    \hspace{0.03\linewidth} 
    \begin{subfigure}[t]{0.68\linewidth} 
        \centering
        \includegraphics[width=\linewidth,keepaspectratio]{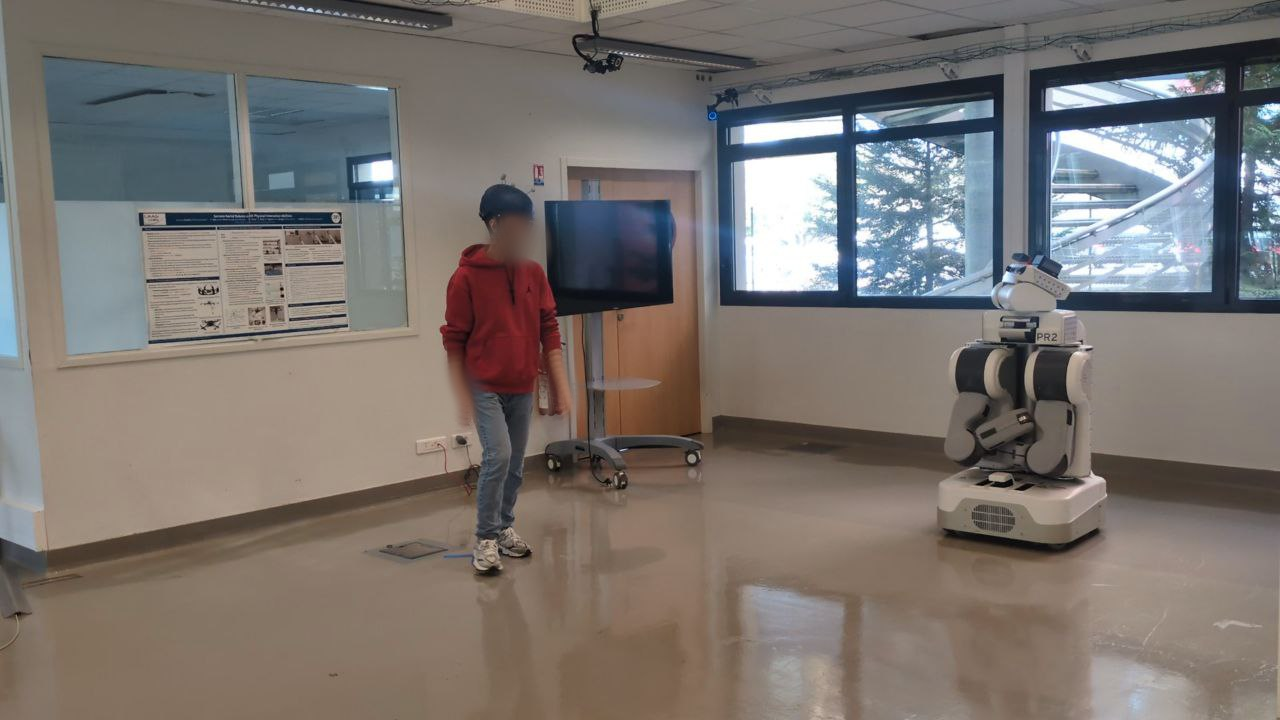}
        \subcaption{Interaction in the real world.}
        \label{fig:rw_image}
    \end{subfigure}

    \caption{Depiction of human interactions in VR and real world setups.
(a) User with VR headset and controller, (b) interaction inside VR along with visual feed used by user for navigation. (c) user wearing motion capture helmet in real world, and (d) real world interaction snapshot.}
    \label{fig:rw_and_vr_image}
\end{figure}

\noindent The study was conducted in two settings:
\vspace{-0.2cm}
\begin{itemize}
    \item \textbf{Real World (RW)}: Experiments took place in a motion-capture equipped arena. Participants wore a head-pose tracking helmet (Fig. \ref{fig:helmet_and_body_markers}) to provide pose data, replicating the information provided by the Sensors and Interface component of the VR framework. The robot localizes itself with its in-built LiDAR in real world, and uses the pose of the human captured with the motion capture system to localize human in the environment.
    \item \textbf{Virtual Reality (VR)}: A replica of the real-world arena was created in VR. Participants wore a VR headset and held a controller (Fig. \ref{fig:user_with_vr_interface}) to control their avatar that navigates in the virtual environment. In this case, the human and robot poses are captured directly from the simulation environment created in Unity.
\end{itemize}

For each scenario and setting, participants navigated between two goal locations (Point A and Point B). Each round consisted of traveling from Point A to Point B and back to Point A. Participants completed ten rounds per scenario in each setting, resulting in a total of 40 navigation rounds per participant.

Participants were informed that the mobile robot would also be navigating independently in the shared space. They were not informed of the robot's goal locations or specific navigation behavior to capture natural human-robot interaction in both RW and VR.

Before starting the VR experiment, participants completed a 3-minute habituation session in VR to familiarize themselves with the control and practice collision avoidance with static obstacles.
The full study duration was approximately 50-60 minutes per participant, including habituation, VR and RW experiments, and questionnaires.

The two scenarios were chosen because they capture diverse interaction dynamics while remaining within the experimental space. They differ in terms of visibility of robot during co-navigation, the navigational choices required to resolve potential conflicts, and available space for navigation decisions. The snapshots in Fig.~\ref{fig:vr_image} and Fig.~\ref{fig:rw_image} illustrate the orthogonal crossing scenario in VR and RW, respectively.

\subsection{Robot in the Scene} 

A PR2\footnote{\url{https://wiki.ros.org/Robots/PR2}} robot was used for the experiments, both in RW and VR. In each scenario, the robot navigated between two goal locations similar to the human participant. Navigation was performed using CoHAN \cite{singamaneni2021human}, a state-of-the-art SRN planner. CoHAN receives both the human's goal and the robot's goal and generates two optimized time elastic bands \cite{rosmann2012trajectory}, one corresponding to the anticipated human trajectory, and one for the robot's own navigation.

In addition to navigation, the robot performed gaze behaviors inspired by the work in \cite{khambhaita2016head}. Using the trajectory produced by CoHAN, the gaze planner anticipates the robot's trajectory approximately one second ahead. When a human enters the robot's field of view within 4 meters, the robot produces a brief saccadic gaze towards the person, simulating socially attentive behavior.

A single planner and robot platform were used across all the experiments to ensure consistency and control. This design allows the study to isolate human responses to the same intelligent, socially aware robot, without introducing variability due to differences in robot morphology, size, or motion dynamics. By maintaining consistent robot behavior, the experiments focus on how humans perceive and react to a socially compliant robot across VR and RW settings.

\subsection{Measurement}
To evaluate the suitability of VR for studying social navigation, participants were randomly assigned to start with either VR or RW to counterbalance potential order effects. After completing the navigational rounds in each setting, participants filled out a questionnaire assessing their perception of the robot's behavior. Table.\ref{tab:questions} shows the items used in the questionnaires. The questionnaire included items on perceived robot's social awareness, the naturalness of motion in VR, and the ease of navigation and control during the task. Responses were recorded on a 5-point Likert scale. From Table.\ref{tab:questions}, Q1 was given after RW, and Q2-Q4 was given after VR.

At the end of the experiment, participants completed an additional questionnaire (Q5-Q6) directly comparing their experiences between VR and RW, covering aspects such as interaction comfort, navigation, and overall system usability. In addition to questionnaire data, the trajectories of both human and robot were recorded to allow detailed analysis of interaction dynamics during the navigation tasks.

\begin{table}[H]
\centering
\begin{tabularx}{\linewidth}{| c | c  | X| }
\hline
\textbf{ID} & \textbf{Condition} & \textbf{Questionnaire statement item} \\
\hline
Q1 & RW  & The robot in RW was socially aware of my presence and navigation. \\

\hline
Q2 & VR  & The robot in VR was socially aware of my presence and navigation. \\
\hline

Q3 & VR & Walking felt natural inside the VR after the habituation session. \\
\hline
Q4 & VR & The VR controls were easy to use for navigation. \\
\hline
Q5 & VR vs RW & Interacting with the robot in VR was as comfortable as in the real world. \\
\hline
Q6 & VR vs RW & The robot’s head behavior in VR matched the real robot’s head behavior. \\
\hline
\end{tabularx}
\caption{Questionnaire statements used in the study. All items were rated on a 5-point Likert scale (1 = Strongly Disagree, 5 = Strongly Agree).}
\label{tab:questions}
\end{table}
\vspace{-1cm}

\section{Results and Discussions} \label{sec:results}
As the questionnaires were different for VR and RW, we could not do a deeper statistical analysis comparing the experiences between the two. However, we report some preliminary impressions of the users in the subsequent subsections. Following these we present quantitative and qualitative analysis that compare the similarities between VR and RW.

\subsection{Perceived Social Awareness of the Robot}
\begin{figure}[htpb]
    \centering
    \begin{subfigure}{0.33\textwidth}
        \includegraphics[width=\linewidth]{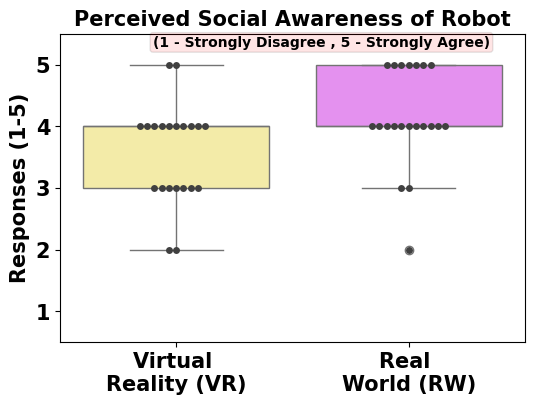}
        \caption{Social Awareness}
        \label{fig:social_awareness}
    \end{subfigure}%
    \begin{subfigure}{0.33\textwidth}
        \includegraphics[width=\linewidth]{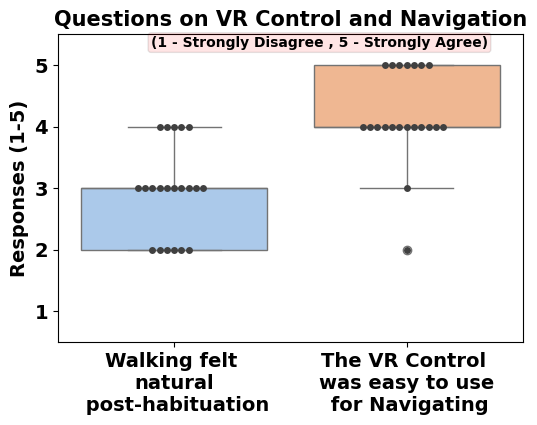}
        \caption{VR Usability\\}
        \label{fig:vr_questions}
    \end{subfigure}%
    \begin{subfigure}{0.33\textwidth}
        \includegraphics[width=\linewidth]{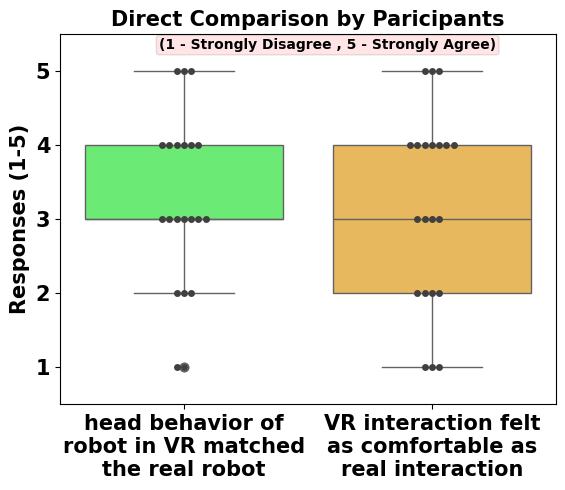}
        \caption{Direct Comparison\\}
        \label{fig:direct_comparison}
    \end{subfigure}

    \caption{Distribution of participant responses to questionnaire items.} 
    \label{fig:placeholder}
\end{figure}

Although robot navigation used the same planner in both RW and VR, participants rated their agreement with statement Q1 and Q2 (Table~\ref{tab:questions}) on a 5-point Likert scale to assess perceived social awareness of the robot in each setting. The distribution of responses is shown in Fig.~\ref{fig:social_awareness}.

While the median ratings of VR and RW were identical (median = 4), individual responses were more varied in VR than in the RW, the correlation between VR and RW ratings was low (r = 0.17). This suggests that while VR can elicit similar tendencies in perception, individual experiences may not consistently match those in the real world.

\subsection{VR Usability}
Participants also evaluated the VR interface. On a 5-point Likert scale (1 =
Strongly Disagree, 5 = Strongly Agree), the perceived naturalness of navigation
in VR was assessed using statement Q3. This item received a median rating of 3 with interquartile region (IQR) of 1.
Participants also responded to Q4 to assess controller usability, which received a higher median rating of 4 (IQR = 1), as shown in Fig.~\ref{fig:vr_questions}.

This indicates that most participants found the VR controls intuitive, although the perceived naturalness of movement within the VR environment was lower.
These findings suggest that while the proposed VR setup enabled effective navigation without major usability issues, improvements could further enhance the naturalness of movement within the environment.

\subsection{Direct Comparison by Participants}
Participants also directly compared their experiences between the VR and RW. They rated their agreement with the statement Q5 to evaluate interaction comfort.
They also responded to the statement Q6 to assess the perceived consistency of robot's head behavior across settings.
Interaction comfort ratings had a median of 3 (IQR = 2), indicating a moderate level of comfort when interacting with the robot in VR compared to RW. This results suggests that the VR setup did not introduce substantial fatigue or discomfort, supporting its feasibility for longer interaction experiments in social navigation contexts.

Similarly, perceived head behavior consistency between VR and RW settings had a median of 3 (IQR = 1), suggesting that participants generally perceived the robot's head movement in VR as moderately consistent with those observed in the real environment. However, this result should be interpreted with caution, as some participants did not notice the robot’s head behavior during their initial interactions and therefore gave lower ratings for Q6.

Overall, the subjective analysis indicates that participants perceived the robot's SRN behaviors similarly in VR and RW settings. Although the naturalness of locomotion in VR was rated slightly lower, participants generally, found the VR controls intuitive and reported moderate interaction comfort compared to the real world experience. 

\vspace{-0.3cm}
\subsection{Navigational Comparisons}
\begin{figure}[h]
    \centering

    \begin{subfigure}{\textwidth}
        \centering
        \textbf{Real World (RW) Navigation}
    \end{subfigure}

    \vspace{0.3em}

    \begin{subfigure}[b]{0.32\textwidth}
        \includegraphics[width=\linewidth]{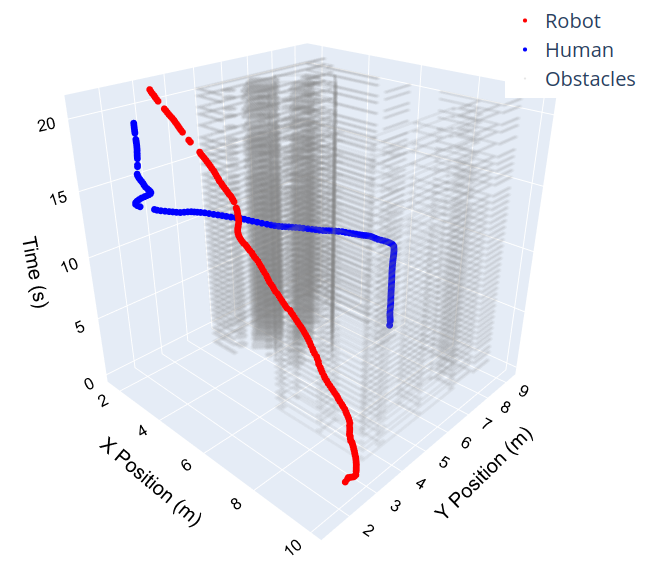}
        \subcaption{Position vs Time}
        \label{fig:ortho_real_3d}
    \end{subfigure}%
    \hfill
    \begin{subfigure}[b]{0.32\textwidth}
        \includegraphics[width=\linewidth]{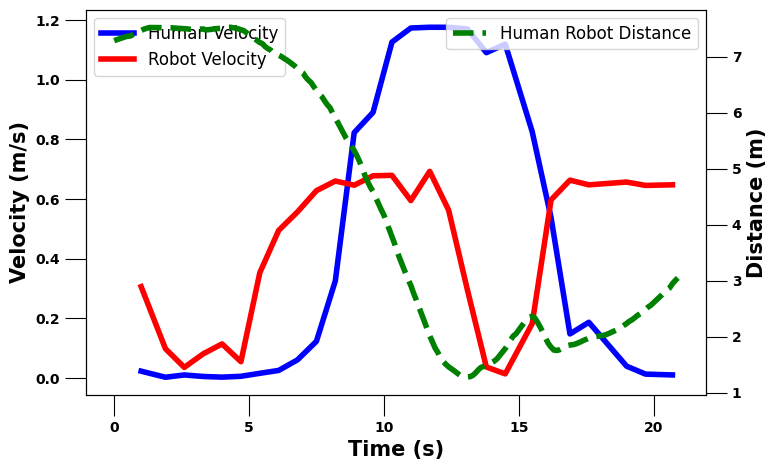}
        \subcaption{Velocity}
        \label{fig:ortho_real_vel_dis}
    \end{subfigure}%
    \hfill
    \begin{subfigure}[b]{0.32\textwidth}
        \includegraphics[width=\linewidth]{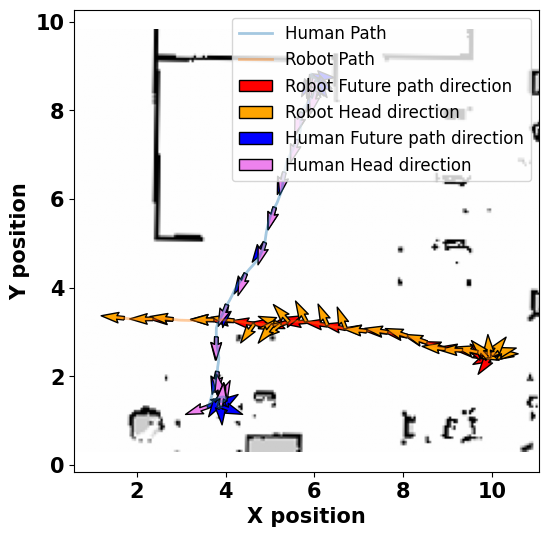}
        \subcaption{Head Behavior}
        \label{fig:head_behavior_a}
    \end{subfigure}

    \vspace{0.8em}

    \begin{subfigure}{\textwidth}
        \centering
        \textbf{Virtual Reality (VR) Navigation}
    \end{subfigure}

    \vspace{0.3em}

    \begin{subfigure}[b]{0.32\textwidth}
        \includegraphics[width=\linewidth]{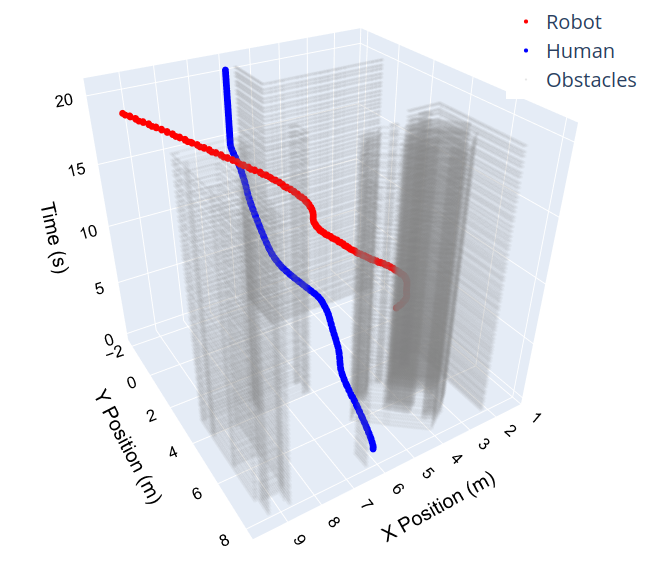}
        \subcaption{Position vs Time}
        \label{fig:ortho_vr_3d}
    \end{subfigure}%
    \hfill
    \begin{subfigure}[b]{0.32\textwidth}
        \includegraphics[width=\linewidth]{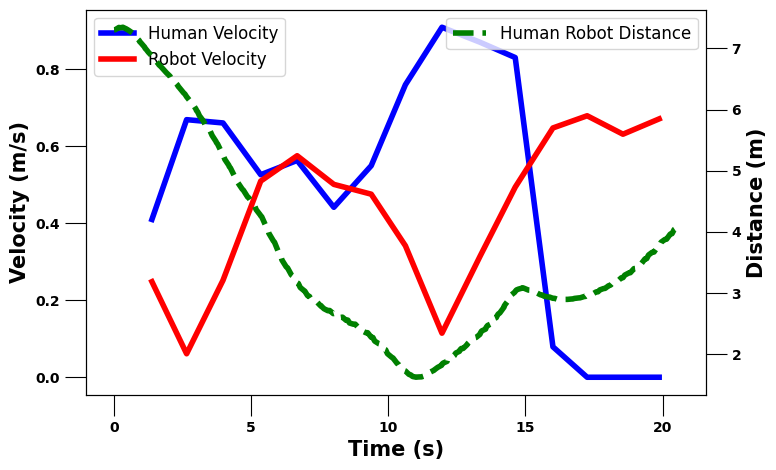}
        \subcaption{Velocity}
        \label{fig:ortho_vr_vel_dis}
    \end{subfigure}%
    \hfill
    \begin{subfigure}[b]{0.32\textwidth}
        \includegraphics[width=\linewidth]{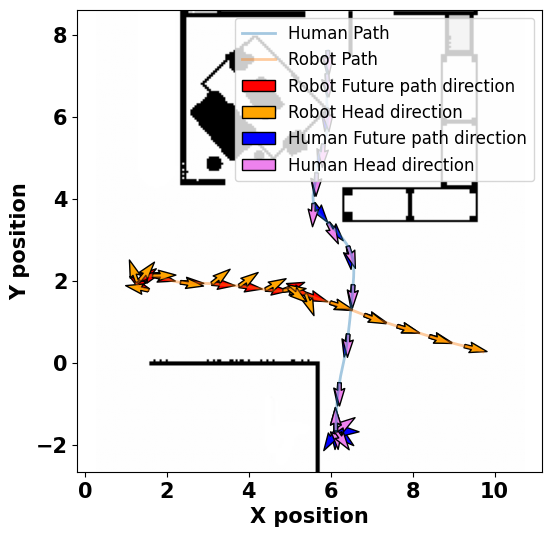}
        \subcaption{Head Behavior}
        \label{fig:head_behavior_b}
    \end{subfigure}

    \caption{Illustration of navigational and head behavior of human and robot. Red: Robot, Blue: Human, Green: Human-Robot Distance, Violet: Head Direction of Human, Yellow: Head Direction of Robot, Black: Static Obstacles.}
    \label{fig:navigation_illustration}
\end{figure}

To analyze the interaction dynamics between the human and the robot, 240 interaction trajectories were selected from the collected dataset (60 interactions for each scenario across both VR and RW settings). Each interaction trajectory corresponds to a navigation episode in which the human and the robot encountered each other while navigating towards their respective goals. Fig.~\ref{fig:navigation_illustration} shows a representative navigation trajectories of participants and robot in both VR and RW for Ortho scenario. The 3D plots (Figs.~\ref{fig:ortho_real_3d}, ~\ref{fig:ortho_vr_3d}) illustrates the positions of both the human (in blue) and robot (in red)  plotted over time, these allow us to see the path and speed modulations of both of them, in this specific case, the speed reduction of the robot can be noticed with the increasing slope of the robot's path with respect to time. The plots in Fig.~\ref{fig:ortho_real_vel_dis} and Fig.~\ref{fig:ortho_vr_vel_dis} show the respective speed modulation of the human and robot and their corresponding human-robot distance during the interaction. We can observe that the velocity profiles are similar for the humans and the robot in RW and VR. This similarity can be observed in Fig.~\ref{fig:head_behavior_a} and Fig.~\ref{fig:head_behavior_b} as well that show the head behaviors of human and robot recorded during the navigation in RW and VR respectively. They illustrate the head direction of the human and robot during navigation, overlaid on the direction of their future navigation.

\vspace{-0.5cm}
\subsubsection{Velocity Profiles}
\begin{table}[h]
\centering
\begin{tabularx}{\textwidth}{|c|>{\centering\arraybackslash}X|>{\centering\arraybackslash}X|>{\centering\arraybackslash}X|>{\centering\arraybackslash}X|>{\centering\arraybackslash}X|}
\hline
\textbf{Scenario - Setting} & \multicolumn{2}{c|}{\textbf{Vel. (Closest)}} & \multicolumn{2}{c|}{\textbf{Jerk (Closest)}} & \textbf{Min. Dist. (H-R)} \\
\cline{2-5}
 & Human & Robot & Human & Robot & \\ 
\hline
Ortho - VR & 0.65 & 0.45 & 0.12 & 0.10 & 1.90 \\
\hline
Ortho - Real & 0.87 & 0.56 & 0.30 & 0.23 & 1.61 \\
\hline
Passby - VR & 0.60 & 0.32 & 0.14 & 0.13 & 1.42\\

\hline
Passby - Real & 1.03 & 0.39 & 0.22 & 0.20 & 1.26\\
\hline
\end{tabularx}
\caption{Average Velocity (Vel.), Jerk of human and robot during their closest interaction, and the distance between them (Min. Dist. (H-R)) during the closest interaction.  }
\label{tab:navigational_comparison_table}
\end{table}
\vspace{-0.1cm}

Table~\ref{tab:navigational_comparison_table} presents the velocity and jerk of both the human and the robot at their closest point of interaction, along with the minimum distance between them during navigation. The velocity and jerk were computed from positions sampled approximately every 0.5 seconds throughout their navigation. The reported values represent the average of each metric across all interaction trajectories for each scenario and setting.

The results indicate that participants moved faster in the real world compared to VR in both scenarios, and maintained slightly larger minimum human-robot distance in VR compared to real world for both scenarios. 
The robot's velocity remains relatively consistent across scenarios and settings.

The smaller jerk values in VR indicate that the velocity modulation was smoother compared to real world setting, where more natural jerk behavior is exhibited.

This suggests that participants maintained a greater safety margin in VR despite similar robot motion profiles. Such differences may arise from perception biases or differences in depth perception and spatial awareness in VR, which should be taken into account in future studies of social navigation.

\vspace{-0.4cm}
\subsubsection{Path Deviations}
\begin{table}[h]
\centering
\begin{tabularx}{\textwidth}{|c|>{\centering\arraybackslash}X|>{\centering\arraybackslash}X|>{\centering\arraybackslash}X|>{\centering\arraybackslash}X|}
\hline

\hline
\textbf{Scenario - Setting} & \multicolumn{2}{c|}{\textbf{Avg. Dev. }} & \multicolumn{2}{c|}{\textbf{Dev. Corr. (w H-R Dist)}} \\
\cline{2-5}
 & Human & Robot & Human & Robot \\ 
\hline
Ortho - VR & 0.25 & 0.19 & -0.26 & -0.37 \\
\hline
Ortho - Real & 0.39 & 0.10 & -0.24 & -0.24 \\
\hline
Passby - VR & 0.69 & 0.24 & -0.42 & -0.25 \\
\hline
Passby - Real & 0.69 & 0.21 & -0.38 & -0.37 \\
\hline
\end{tabularx}
\caption{Average Path Deviation (Avg. Dev.) of Human and Robot from their shortest paths to goals, and their Pearson Correlation Coefficient (Dev. Corr. (w H-R Dist)) with respect to human-robot distance.}
\label{tab:path_deviation}
\end{table}

Further, we compare the path deviation of human and robot from their shortest path to goal. A shortest path (for both human and robot) between the two goal locations in each scenario and setting is derived using Dijkstra on the respective maps of the arena. The trajectories are then used to calculate the deviation defined as the shortest distance from the corresponding shortest path.

Table~\ref{tab:path_deviation} shows the average deviation which indicates the deviation averaged over a single trajectory, and the values indicate an average of this value over all the interaction trajectories. The table also reports the Pearson correlation coefficient of the deviation with respect to the human-robot distance. The results show that participants exhibited larger path deviations in the Passby scenario compared to the Ortho scenario in both VR and RW.

In contrast, the robot’s path deviations remain relatively consistent across scenarios and settings, reflecting the stable behavior of the navigation planner. The negative correlation between the deviations and human-robot distances indicates that participants tended to deviate more from their shortest path when the robot was closer. 
The consistent deviation behavior across VR and RW further supports the suitability of VR for studying social navigation behavior.

\vspace{-0.6cm}
\subsubsection{Human Head Behavior}
\begin{table}[h]
\centering
\begin{tabularx}{\textwidth}{|c|>{\centering\arraybackslash}X|>{\centering\arraybackslash}X|>{\centering\arraybackslash}X|>{\centering\arraybackslash}X|>{\centering\arraybackslash}X|}

\hline
\textbf{Scenario - Setting} & \multicolumn{3}{c|}{\textbf{Towards Future Nav.}} & \multicolumn{2}{c|}{\textbf{Towards the Robot}} \\
\cline{2-6}
 & Cos. Dis.(WA) & Corr. (w H-R Dis.)(WA) & Cos. Dis.(AC) &  Cos. Dis. (WA) & Corr. (w H-R Dis.) (WA)\\ 
\hline
Ortho - VR & 0.76 & -0.20 & 0.60  & 0.49 & 0.10\\
\hline
Ortho - Real & 0.76 & -0.07 & 0.58 & 0.54 & 0.22\\
\hline
Passby - VR &  0.70  & -0.34 & 0.58 & 0.59 & -0.35 \\
\hline
Passby - Real & 0.78  & -0.40 & 0.52 & 0.65 & -0.10\\
\hline
\end{tabularx}
\caption{Head Behavior Towards Future Navigation and Towards Robot, Cosine Distances (Cos. Dis.), and their Pearson Correlation Coefficient with Human-Robot Distance (H-R Dis.) while approaching (WA), and after crossing (AC).} 
\label{tab:human_head_behavior}
\end{table}
\vspace{-0.1cm}

Table~\ref{tab:human_head_behavior} shows the human's head behavior in different scenarios and settings from the interaction trajectories. 
The table indicates cosine distance between the human's head direction during the navigation, towards their own future navigation, and towards the robot, thresholded based on the field of view of the human. Due to the limited peripheral field of view of the VR headset, the field of view in VR is assumed to be \(90^\circ\), and we assume the field of view of humans during RW to be \(150^\circ\). When the vector lies outside the field of view, the cosine distance is set to zero.

To analyze head behavior across different stages of the interaction, each trajectory is divided into two segments: While Approaching (WA), corresponding to the period before the human and robot cross each other, and After Crossing (AC), corresponding to the period after the crossing event. The table also indicates the Pearson Correlation Coefficient of the respective cosine distances with respect to the human-robot distance in WA. 

The results show that participants' head were mostly aligned with their future navigation while approaching the robot, in all the scenarios and settings, with a weak negative correlation with the human robot distance. In contrast, after crossing, the alignment with the future navigation direction decreases. Overall, the head behavior towards the future navigation remains consistent across VR and RW.

Simultaneously, during WA, the robot was moderately present in the field of view of the human during navigation. Given that the cosine distances were thresholded, the similar values of the average cosine distance could indicate participants exhibiting a head behavior to have the robot in their field of view equally in both VR and RW. Although the correlation with human-robot distance is not consistent across the scenarios, it remains negative in passby, and positive in ortho, irrespective of the setting. This may indicate that humans adapt their visual attention to maintain awareness of dynamic obstacles depending on the potential interaction with them.

Overall, the similarities in head orientation patterns across VR and real world settings suggest that VR is capable of capturing key multimodal behaviors during human-robot navigation.

%% file: conclusion.tex
\vspace{-0.1cm}
\section{Limitations}
\vspace{-0.1cm}
\label{sec:limitations}
Although the work presents some early evidence of the suitability of VR for multimodal SRN, there are some limitations of the proposed VR framework.
\vspace{-0.1cm}
\begin{itemize}
    
    \item \textbf{Absence of auditory cues:}
    Participants noted that subtle robot-generated sounds in the real world help anticipate the robot's motion and intentions. Without these cues, they relied more heavily on visual monitoring. Future iterations could integrate spatialized audio to enhance perceptual realism and support richer multimodal interaction. In addition, the framework could record verbal reactions from human user, enabling the capture of spoken feedback or instruction during the interaction,

    \item \textbf{Limited multimodal capture:}
    While VR enables the capture of multiple modalities, the current framework records only planar navigation and head orientation. Future versions could incorporate additional motion cues, such as full 3D human pose (e.g., via localized body suits) and direct gaze tracking using dedicated eye-tracking sensors. Because  the environment is simulated in 3D, it would also be possible to compute perceptual information, such as visibility relationships between the human and the robot, including occlusions caused by objects, windows, or walls, as well higher-level semantic information about the environment. 
    
    \item \textbf{Restricted field of view:}
    The VR headset’s limited peripheral view occasionally prevented participants from noticing the robot as easily as they might in the real world. Although this is primarily a hardware limitation, future work could explore software adjustments or alternative hardware to mitigate this issue.

    \item \textbf{Limited diversity of participants:}
    The current study involved homogeneous population. Future experiments could include broader user group, such as children or elderly participants, and explore the use of different human avatar (e.g., representing a child or an older person) to investigate how human characteristics influence robot navigation behavior and interaction dynamics.

    \item \textbf{Reduced Naturalness of Navigation:}
    As discussed earlier, participants reported reduced naturalness during navigation inside the VR environment. Future iterations could incorporate treadmills that are specifically designed for virtual reality, to enhance the naturalness of navigating inside VR.

\end{itemize}

Despite these limitations, the proposed VR framework already provides a significantly richer and more controllable experimental setting than many existing approaches. It enables the collection of detailed multimodal interaction data that can support more informative learning processes and offers a promising basis for reproducible benchmarking of human-robot navigation methods.

While the current analysis focuses on descriptive and correlational trends, preliminary observations (e.g., similar medians and consistent patterns across velocity, deviation, and head behavior metrics) indicate convergence between VR and real-world interaction dynamics. These results motivate deeper statistical validation in future studies using larger sample sizes and more extensive datasets.

\vspace{-0.1cm}
\section{Conclusions and Future Works}
\label{sec:conclusion}
\vspace{-0.1cm}
Although the scenarios studied in this work cover key interaction patterns, future studies could incorporate more complex environments and obstacle configurations to study human navigational strategies in richer interaction contexts.

While other robots can be simulated within the proposed VR framework, future work will explore additional robot embodiments and alternative planners to examine the generalizability of human responses and the influence of different navigation strategies on social interaction.

Overall, the convergence of subjective ratings, trajectory metrics, and head orientation patterns supports VR as a valid platform for studying multimodal socially-aware navigation. Although minor differences in perceived naturalness and perception biases remain, the key social interaction dynamics between humans and robots are largely preserved. 
These results indicate that VR provides controlled, reproducible, and flexible environment for investigating human-robot co-navigation and developing multimodal SRN strategies.

While SRN planners will continue to benefit from refinement using real-world interaction data, VR could provide a cost-efficient platform for collecting rich and near-natural human behavior. Furthermore, the proposed framework shows strong potential for future extensions toward studying more comprehensive embodied interactions, integrating additional human modalities and higher-dimensional motion cues. Future iterations of the VR framework could integrate localized sounds of the environment, treadmills specifically designed for VR interactions, and capture whole body pose of humans with motion capture body suits.

\section*{Acknowledgments}
This work was supported by Horizon Europe under the MSCA grant agreement No 101072488 (TRAIL) and by the French National Research Agency (ANR) under the OSTENSIVE project ANR-24-CE33-6907-01. The Authors thank Shriram Hari for software support during the development of the VR framework, and all the participants for their participation in the study.